\documentclass{article}

 \usepackage{caption}

\usepackage[final,nonatbib]{neurips_2025_ml4ps}
\usepackage{verbatim}
\usepackage{multirow}
\usepackage{float} 
\usepackage[utf8]{inputenc} % allow utf-8 input
\usepackage[T1]{fontenc}    % use 8-bit T1 fonts
\usepackage{hyperref}       % hyperlinks
\usepackage{url}            % simple URL typesetting
\usepackage{booktabs}       % professional-quality tables
\usepackage{graphicx}
\usepackage{amsfonts}       % blackboard math symbols
\usepackage{nicefrac}       % compact symbols for 1/2, etc.
\usepackage{microtype}      % microtypography
\usepackage{xcolor} 
 
\def\R{\mathbb R}
\def\C{\mathbb C}

\def\E{\mathbb E}

\newcommand*{\norm}[1]{\left\|#1\right\|}

\newcommand*{\card}[1]{\left|#1\right|}

\def\R{\mathbb R} 

\newcommand{\mbf}{\mathbf}

\newcommand{\mcl}{\mathcal}
 
\usepackage{amssymb}
\usepackage{amsfonts} 
\usepackage{amsmath} 
\usepackage{url} 
\usepackage{amsthm}
\usepackage{algorithm}
\usepackage{algpseudocode}
\usepackage{multicol}
\usepackage{bm}  % for bold math
\usepackage{enumitem}
\usepackage{authblk}

\title{FlowTIE: Flow-based Transport of Intensity Equation for Phase Gradient Estimation from 4D-STEM Data}
\author[1]{Arya Bangun$^{*}$}
\author[2,3,4]{Maximilian Töllner$^{*}$}
\author[1]{Xuan Zhao$^1$\authorcr} 
\author[2,3,4]{Christian Kübel}
\author[1]{Hanno Scharr}

\affil[1]{IAS-8, Forschungszentrum J\"ulich, Germany}
\affil[2]{Institute of Nanotechnology, KIT, Germany}
\affil[3]{ Karlsruhe Nano Micro Facility (KNMF), KIT}
\affil[4]{Department of Materials Science, TU Darmstadt}
\affil[1]{\texttt{\{a.bangun, xu.zhao, h.scharr\}@fz-juelich.de}}
\affil[2]{\texttt{\{maximilian.toellner, christian.kuebel\}@kit.edu}}

\begin{document}
\maketitle
\begin{abstract}
We introduce FlowTIE, a neural-network-based framework for phase reconstruction from 4D-Scanning Transmission Electron Microscopy (STEM) data, which integrates the Transport of Intensity Equation (TIE) with a flow-based representation of the phase gradient.  This formulation allows the model to bridge data-driven learning with physics-based priors, improving robustness under dynamical scattering conditions for thick specimen. The validation on simulated datasets of crystalline materials, benchmarking to classical TIE and gradient-based optimization methods are presented. The results demonstrate that FlowTIE improves phase reconstruction accuracy, fast, and can be integrated with a thick specimen model, namely multislice method. 
\end{abstract}
{\let\thefootnote\relax\footnote{\hspace{-0.6cm}{$^*$Equal contribution}}} 
\section{Introduction}
Electron microscopy (EM) has become an indispensable tool in modern scientific research due to its ability to resolve structures at nanometer to sub-nanometer scales. %Unlike conventional light microscopy, which is limited by the wavelength of visible light, EM utilizes accelerated electrons with much shorter wavelengths, enabling visualization of ultrastructural details with exceptional resolution and contrast. 
In (scanning) transmission electron microscopy (STEM), the reconstruction of the phase of the electron wave after interaction with an object under investigation is essential for understanding a wide range of physical phenomena, for instance electromagnetic fields and compositional variations at the atomic scale \cite{MuellerCaspary.2017}. Both are critical for the development of advanced materials, e.g. for quantum devices and magnetic storage technologies. However, direct measurement of the phase is not feasible, making phase retrieval a central challenge in electron microscopy. Several techniques have been developed to address this limitation. Iterative ptychography approaches e.g., \cite{Hoppe.1969, Thibault.2008, Thibault.2009, Pfeiffer.2018}, are computationally intensive and not compatible with large-angle Lorentz 4D-STEM imaging \cite{Hale.1959, Kang2025}. Deterministic methods such as differential phase contrast (DPC\cite{Rose.1977, Chapman.1978, Haider.1994}/iDPC\cite{Lazic.2016} or COM\cite{MuellerCaspary.2017}/iCOM\cite{Lazic.2016}), and Transport of Intensity Equation (TIE)\cite{Mitome.2021,Teague.1983} provide faster, more direct solutions with analytical formulation but are typically limited in resolution or applicability, especially under non-ideal imaging conditions. %Iterative methods, such as iterative ptychography are computationally intensive and {difficult to ensure } compatibility with {large-angle } Lorentz {4D-STEM}  imaging \cite{Hale.1959, Kang2025}, while deterministic methods like DPC and TIE \cite{Phatak.2016, teague1982irradiance,teague1983deterministic} are faster but limited in resolution or require restrictive sample conditions. 
%{\let\thefootnote\relax\footnote{\textsuperscript{*}Shared first authorship.}}
%{\let\thefootnote\relax\footnote{{Code: \url{https://jugit.fz-juelich.de/ias-8/r3dm/}}}} 
Recently, the development of data-driven methods has accelerated rapidly, driven by the increasing availability of large-scale datasets and advancements in machine learning. These approaches offer flexible, scalable alternatives to model-based techniques, particularly in domains where physical modeling is complex or poorly understood. One prominent example is generative models such as diffusion models and flow-matching \cite{ho2020denoising, songscore, lipmanflow}, which have demonstrated remarkable success in generating high-fidelity data across domains, such as natural images, molecular structures, and material generation \cite{songsolving, hoogeboom2022equivariant, zeni2023mattergen}. While such methods have shown compelling results in image synthesis, their application to high-resolution scientific imaging, particularly 4D-STEM, remains underexplored, due to its high dimensionality, noise sensitivity, and dependence on precise physical interpretation. As a result, there is still a significant gap in both investigating and adapting generative model in 4D-STEM. We propose an integration of the Transport of Intensity Equation (TIE) and flow matching, coined as FlowTIE. FlowTIE offers a fast, analytical solution to the phase retrieval problem, hence, avoiding the intensive computational demands by iterative algorithms.

\section{Methodology}
\begin{figure}[H]
    \centering
\includegraphics[width=0.95\linewidth]{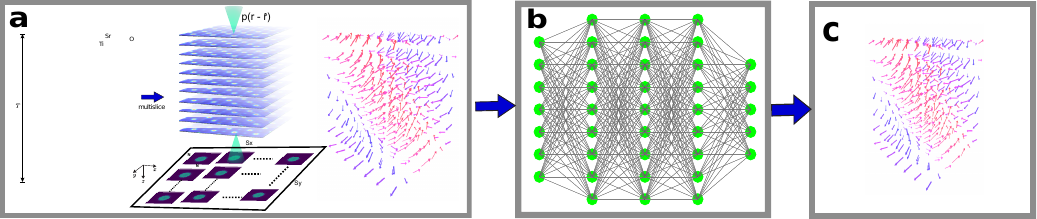}
    \caption{ 
4D-STEM diffraction data and vector field of the phase are simulated from multislice models, which serve as training data for the neural network model
shown in (b). After training, the model can generate vector fields based on the intensity distribution of diffraction patterns (c).}
    \label{fig:stem_setting}
\end{figure}

\paragraph{Notation}
Vectors  $\mathbf{x} \in \C^L$ and matrices  $\mathbf{A} \in \C^{K \times L}$ are written in bold small-cap and bold big-cap letter, respectively. $\mathcal{F}_{\mbf r}$ is two-dimensional Fourier transform applied to spatial grid. i.e., real space in TEM terminology. The conjugate transpose is written as $\mbf{A}^H$. The absolute value and square root is applied element-wise to vectors or matrices.
\paragraph{Multislice Model} The multislice method is a common computational approach to model dynamical scattering effects inside of thick crystalline materials \cite{cowley1957scattering}. The incoming electron wave is sequentially propagated through slices, i.e., z$-$axis, accounting for both phase shifts due to atomic potentials and scattering effects.  Given the first slice of crystal on coordinate $\mbf r = (x,y)$, $O_1\left(\mbf r\right) = e^{i\sigma V_z\left(\mbf r\right)}$, where $V_z\left(\mbf r\right) = V_z\left(x,y\right) =   \int_{z}^{z + \Delta_z} V\left(x,y,z\right) \mathrm{d}z$ is projected potential of crystalline materials with interaction constant $\sigma$. The interaction between the incoming focused electron wave with raster scan $P\left(\mbf r - \mbf{\hat r}\right)$ at a  specific scanning point coordinate $\mbf{\hat r}$ is given by the product $$E_1(\mbf r, \mbf{\hat r}) = O_1\left(\mbf r\right) P\left(\mbf r - \mbf{\hat r}\right),$$ where $E_1$ is the exit wave from the first slice and it is propagated with propagation model $\mcl{V}$, i.e., Fresnel propagation, before interacting with the next slice and producing exitwave $$E_2\left(\mbf r, \mbf{\hat r}\right) = O_2\left(\mbf r\right) \mcl{V}\left(E_1\left(\mbf r, \mbf{\hat r}\right) \right).$$ This process is repeated until $m$ slices have been traversed by the electron wave, resulting in a final exit wavefunction.  The intensity recorded at 4D-STEM detector can be written as $$
I \left(\mbf q, \mbf{\hat r}\right) = \card{\mcl{F}_{\mbf r}\left( E_m\left(\mbf r, \mbf{\hat r}\right) \right) }^2.$$ In total we have total scan points $S_y \times S_x$ and detector dimension $N \times N$.  Fourier transform on coordinate $\mbf r$, $\mcl F_{\mbf{r}}$,  maps the exitwave in the far-field and we record intensity in Fourier space $\mbf q$, or reciprocal space in TEM terminology. The scan coordinates $\mbf{\hat r}$ are still in  real space. Figure \ref{fig:stem_setting} visualizes the 4D-STEM acquisition setting.
Apart from the classical representation of multislice formalism, one can write the scattering process as a matrix transformation to disentangle probe effect, as discussed in \cite{bangun2022inverse}. Hence, for all scan points $S = S_y \times S_x$ and detector dimension $N \times N$,  the 4D-STEM diffraction patterns can be written as
$$\mbf{I} = \card{\mbf E}^2=\card{\mbf{F}_{2D} \mbf A \mbf P}^2 \in \R^{N^2 \times S},$$
where the $\mbf{F}_{2D}$ is the matrix representation of two-dimensional Fourier transform, $\mbf A \in \C^{N^2 \times N^2}$ is the matrix potential and $\mbf P \in \C^{N^2 \times S}$ is the matrix representation of the focused electron wave for all scan points. Regardless of whether the classical multislice or matrix model is used, the key challenge in 4D-STEM multislice is to estimate the (gradient) phase, i.e., potential of crystalline materials $V(x,y,z)$, given the intensity measurements. 
\paragraph{Transport of Intensity Equation (TIE)} 
The relationship  between intensity $I$ variations along the beam direction $z-$axis and the phase $\phi$ of the electron wave can be described with the continuity equation, due to the principle of conservation of a physical quantity, i.e., intensity, derived in \cite{teague1982irradiance, teague1983deterministic}
\begin{equation}
\label{eq:cont_eq}
\frac{\partial I\left(x,y,z\right)}{\partial z } = - \frac{\lambda}{2\pi}\nabla_{x,y} \cdot \left(I\left(x,y,z\right) \nabla_{x,y} \phi \left( x,y, z\right)\right),
\end{equation}
where the $\lambda$ is the electron wavelength.  In most cases, the constant transverse intensity assumption, i.e., thin specimen \cite{bostan2014phase}, is given to simplify the model as Poisson equation:
\begin{equation}
\label{eq:poiss_solver}
\frac{\partial I\left(x,y,z\right)}{\partial z } \approx - \frac{\lambda}{2\pi}\nabla_{x,y}^2 \phi\left(   x,y, z\right),
\end{equation}
\cite{paganin1998noninterferometric,allen2001phase, roddier1991wavefront, gureyev1997rapid}, or the summary paper in \cite{zuo2020transport}. The key idea to solve \eqref{eq:poiss_solver} is by employing the property of Fourier transform of Laplacian and estimating the phase in the Fourier space.  
Leveraging the property of the continuity equation \eqref{eq:cont_eq}, we can develop a machine learning framework to estimate the vector field $\mbf{v}_{gt} = \nabla_{x,y} \left( \phi \left(x,y,z\right)\right)$, similar to flow matching approach in generative model machine learning. Apart from the possibility to generalize the model for any measurement data, incorporating neural network model to estimate the vector field, i.e., phase gradient, and solving the continuity equation does not require the thin specimen assumption.
%\paragraph{Flow Matching}
%In the generative model framework, flow matching refers to aligning the generated data distribution with the true data distribution by adjusting the flow of information through a model \cite{lipmanflow}. The goal is to ensure that the flow through the generative model matches the underlying data distribution. The initial idea is that the flow from one distribution to another distribution should fulfill the continuity equation in terms of the probability path.  
%Given a vector field $v: \mathbb{R}^d \times [0,1] \to \mathbb{R}^d$, the continuity equation in terms of  
%the probability path $p_t(x)$  is given by
%\begin{equation}
%\frac{\partial p_t}{\partial t} + \nabla \cdot (p_t v_t) = 0.
%\end{equation}
%In the training, the goal is to minimize the expected value loss between the neural network model $u_\theta(x_t, t) $ and the ground truth vector field from the data distribution $v_t(x)$, namely
%$$
%\underset{\theta}{\text{min}} \,\E_{t, p_t(x)} \norm{u_\theta(x_t, t) - v_t(x)}_2^2.
%$$
%After estimating the vector field, in the inference process an ordinary differential equation (ODE) solver is used to generate the probability distribution with respect to the time step $t$. This ensures that the probability mass is conserved along the flow. Since the vector field $v$ governs the probability path for the specific time $t$, estimating the vector field can be used to generate the probability distribution for an arbitrary time step $t$. This is the key idea of the generative model using flow matching.  
\paragraph{FlowTIE} Suppose we acquire series of intensity from focused and defocused 4D-STEM data over finite difference along $z-$axis, with a series of defocus values $-\Delta_{z}, 0, \Delta_{z}$ written as $I_{\Delta_z}\left(\mbf q, \mbf{\hat r}\right)$. To approximate the derivative, we use the finite difference method by incorporating triplets, $\frac{\partial I_z\left(\mbf q, \mbf R\right)}{\partial z} \approx I_{2\Delta_z}\left(\mbf q, \mbf{\hat r}\right) = \frac{I_{\Delta_z}\left(\mbf q, \mbf{\hat r}\right) - I_{-\Delta_z}\left(\mbf q,  \mbf{\hat r}\right)}{2\Delta_z}$, where $I_0\left(\mbf q, \mbf{\hat r}\right)$ is the intensity acquired from focused probe. Additionally, we
have neural network $u_\theta$, where $\theta$ represents the neural network parameters. FlowTIE learns vector fields of phase distributions from exit waves of 4D-STEM data by minimizing the mean squared error loss
$$\mcl{L}_{\text{vf}}\left(\theta\right) = \E\norm{u_{\theta}\left(I_{2\Delta_z}\left(\mbf q, \mbf{\hat r}\right) \right) - \nabla_{\mbf{\hat r}} \left( \phi \left(\mbf q, \mbf{\hat r}\right)\right)}_2^2$$ and continuity loss given as $$
\mcl{L}_{\text{cont}} \left(\theta \right)= \E\norm{I_{2\Delta_z}\left(\mbf q, \mbf{\hat r}\right) + \frac{\lambda}{2\pi} \nabla_{\mbf{\hat r}}\left(I_0\left(\mbf q, \mbf{\hat r}\right) u_{\theta}\left(I_{2\Delta_z}\left(\mbf q, \mbf{\hat r}\right) \right)\right)}_2^2.$$
Similarly, to control the prediction of the phase, we incorporate an integrator model to estimate the phase $\phi\left(\mbf q, \mbf{\hat r}\right)$, for instance, by applying a Fourier transformation after taking the derivative of the vector field to generate the Laplacian. The phase loss can be written as
$$\mcl{L}_{\text{phase}}\left(\theta \right) = \E\norm{\texttt{integrator}\left(u_{\theta}\left(I_{2\Delta_z}\left(\mbf q, \mbf{\hat r}\right) \right)\right) - \phi\left(\mbf q, \mbf{\hat r}\right)}_2^2.$$
Therefore, the total training loss for FlowTIE is given as follows
\begin{equation}
\label{eq:loss_total}
\mcl{L}_{\text{total}}\left(\theta \right) = \alpha \mcl{L}_{\text{vf}} \left(\theta \right)+ \beta \mcl{L}_{\text{cont}} \left(\theta \right)+ \gamma \mcl{L}_{\text{phase}}\left(\theta \right),    
\end{equation}
where $\alpha,\beta, \gamma$ are the weighting factors for each loss. The algorithm for training and inference can therefore be represented as shown in Algorithm 
\eqref{algo:flowTIE_train} and Algorithm 
\eqref{algo:flowTIE_infer}.
\begin{minipage}[t]{0.48\textwidth}
\begin{algorithm}[H]
\caption{Training FlowTIE}
\label{algo:flowTIE_train}
\begin{algorithmic}[1]
\State \textbf{Initialization:} 
\begin{itemize}[noitemsep,topsep=0pt,left=0pt]
    \item Intensity over  crystalline materials $I_{2\Delta_z}\left(\mbf q, \mbf{\hat r}\right) $, neural network model $u_{\theta}$
    \item Epochs, ground truth $\phi_{\text{gt}} \left(\mbf q, \mbf{\hat r}\right)$ and vector field $\nabla_{\mbf{\hat r}} \left( \phi \left(\mbf q, \mbf{\hat r}\right)\right)$, weighted loss $\alpha. \beta, \gamma$
\end{itemize}    
\For{each epoch}
\For{each batch data}
    \State Optimize $u_{\theta}$ in \eqref{eq:loss_total}, \makebox[0pt][l]{\(\min_\theta \mathcal{L}_{\text{total}}(\theta)\)}
\EndFor
\EndFor
\State \Return Learned neural network $u_{\theta^*}$
\end{algorithmic}
\end{algorithm}
\end{minipage}
\hfill
\begin{minipage}[t]{0.48\textwidth}
\begin{algorithm}[H]
\caption{Inference FlowTIE}
\label{algo:flowTIE_infer}
\begin{algorithmic}[1]
\State \textbf{Input:} $I_0(\mbf{q}, \mbf{\hat{r}})$, $I_{2\Delta_z}(\mbf{q}, \mbf{\hat{r}})$, $u_{\theta^*}$, probe $p(\mbf{r} - \mbf{\hat{r}})$
\State Predict vector field of exit wave: 
$\bm{v}_{\text{pred}} = u_{\theta^*}(I_{2\Delta_z})$
\State Reconstruct phase of exit wave: 
$\phi_{\text{pred}}(\mbf{q}, \mbf{\hat{r}})$
\State Construct exit wave: \\\makebox[0pt][l]{$
E_{\text{pred}}\left(\mbf q, \mbf{\hat r}\right) = \sqrt{I_0(\mbf{q}, \mbf{\hat{r}})} \cdot e^{j\phi_{\text{pred}}(\mbf{q}, \mbf{\hat{r}})}$}
\State Convert to matrix: 
$\mbf{E}_{\text{pred}}, \mbf{P} \in \mathbb{C}^{N^2 \times S}$
\State Est. matrix potential:\makebox[0pt][l]{
$\mbf{A}_{\text{pred}}   \approx \mbf{F}_{2D}^{-1} \mbf{E}_{\text{pred}}  \mbf{P}^H$}
\State Project phase:
$\phi_{\text{proj}} \in \mathbb{R}^{S_y \times S_y}$
\State \textbf{Output:} $\mbf{A}_{\text{pred}}, \phi_{\text{proj}}$
\end{algorithmic}
\end{algorithm}
\end{minipage}

\section{Numerical Results}
\begin{figure}[H]
    \centering
\includegraphics[width=0.7\linewidth]{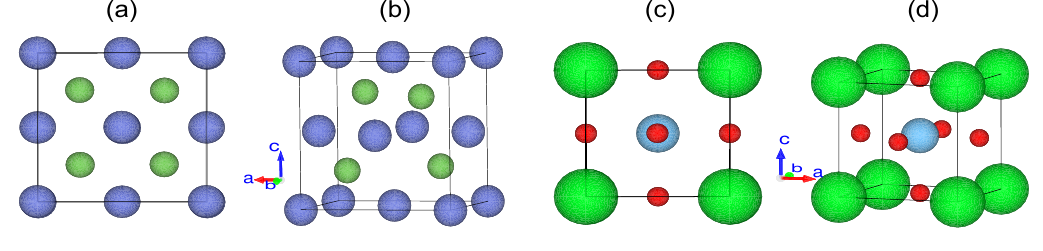}
    \caption{Material used for test data, (a) Gallium Arsenide (GaAs) projection over $z-$ axis, (b) 3D view of GaAs, (c) Strontium Titanate (SrTiO$_3$) projection over $z-$ axis, (d) 3D view of SrTiO$_3$. The unit cell dimension for both SrTiO$_3$ and GaAs are $(a,b,c)= ( 5.6533,5.6533,5.6533)$ and $(a,b,c)= (3.905, 3.905, 3.905)$ all units are in Å, respectively }
\label{fig:materials}
\end{figure}

\begin{figure}[htb]
    \centering \includegraphics[width=1\linewidth]{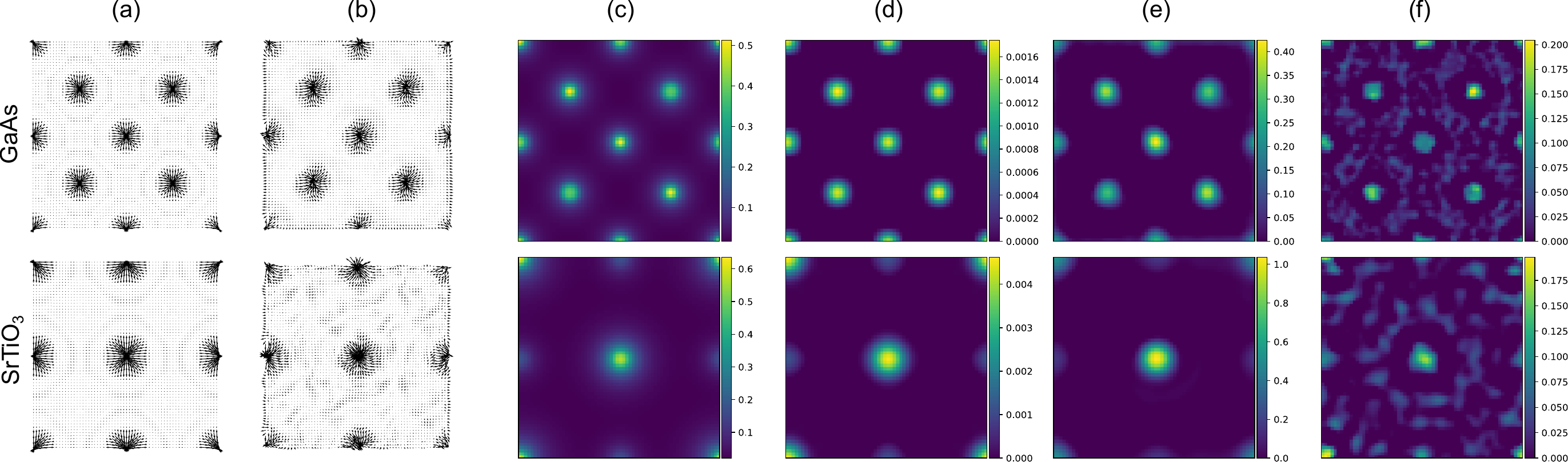}    \caption{Reconstruction result (a) Ground truth of phase gradient, (b) FlowTIE, (c) Ground truth of projected phase, (d) TIE, (e) FlowTIE, (f) Gradient Descent.}\label{fig:gradient_project_phase}
\end{figure}

\paragraph{Dataset and Model Architecture} The cubic crystal system for training data is generated from Materials Project dataset \cite{jain2013commentary}, where the energy potential is computed using Kirkland’s model \cite{kirkland1998advanced}. Ground truth slices of projected potential and phase gradient are produced using the multislice algorithm from \cite{durham2022accurate} with pixel dimension $N \times N = 64 \times 64$. For training, we randomly select 100 structures, with a 0.9/0.1 split between training and validation. For testing, we evaluate the model on well-known materials: GaAs and SrTiO$_3$, as shown in Figure \ref{fig:materials}, where the intensity is generated with aberration-free probe with total scan $S_y \times S_x = 64 \times 64$. Our \texttt{FlowPredictorModel} is a lightweight 4-layer convolutional encoder-decoder network for predicting dense 2D flow fields across input channels. It is designed for multi-channel spatial data where each channel requires a separate 2D flow vector. The encoder (\texttt{ConvEncoder}) uses two convolutional layers with batch normalization and \texttt{GELU} activation to convert input tensors of shape $(B, N^2, S_y, S_x)$ into feature maps of shape $(B, D, S_y, S_x)$, where $D = 128$. The decoder applies two more convolutional layers to output tensors of shape $(B, 2, N^2, S_y, S_x)$, representing horizontal and vertical flow components. Training is performed for 20,000 epochs using the \texttt{AdamW} optimizer with a learning rate of $10^{-4}$. All loss terms are equally weighted ($\alpha = \beta = \gamma = 1$), and no hyperparameter tuning is applied.
\paragraph{Description of the comparative methods}
To assess the performance of the reconstruction algorithm, we compare FlowTIE with the classical TIE solved via the Poisson equation (Fourier approach) and with gradient descent that estimates the potential matrix by directly minimizing
$\underset{\mbf A}{\text{min}}\,\norm{\sqrt{\mbf{I}} - \card{\mbf{F}_{2D}\mbf{A}\mbf{P}}}_F^2$
with respect to $\mbf A$. In this case, the probe $\mbf P$ is initialized as an aberration-free point-spread function, and the optimization is run for 100 iterations.
\paragraph{Estimation of Projected Phase and Gradient Phase}
Figure \ref{fig:gradient_project_phase} shows the phase vector field estimated by FlowTIE for crystalline materials, as well as the projected phase. The model successfully predicts consistent vector fields and projected phase for our test materials, GaAs and SrTiO$_3$. While minor artifacts are present, the overall direction and structure of the vector fields are clearly captured, supporting the use of FlowTIE for vector-field estimation. The dynamic range of the reconstructed projected phase (in radians) closely matches the ground truth.  

In contrast, the Gradient Descent baseline reveals atomic structure but exhibits pronounced artifacts. No explicit regularization was used in that optimization, which likely explains why vanilla gradient descent fails to recover a clean reconstruction.
% --- Table 1: MSE by thickness ---
\begin{table}[H]
\centering
\captionsetup{skip=10pt}
\begin{tabular}{lcccc}
\toprule
\textbf{Material} & \textbf{Thickness (Å)} & \textbf{MSE (TIE)} & \textbf{MSE (FlowTIE)} & \textbf{MSE (Gradient Descent)} \\
\midrule
GaAs                 & 5.7  & 0.0075 & 0.0038 & 0.0068 \\
GaAs                 & 28.3 & 0.1485 & 0.1167 & 0.1482 \\
SrTiO\textsubscript{3} & 3.9  & 0.0073 & 0.0075 & 0.0072 \\
SrTiO\textsubscript{3} & 19.5 & 0.1520 & 0.0853 & 0.1517 \\
\bottomrule
\end{tabular}
\caption{Phase estimation error (MSE) at different thicknesses for GaAs and SrTiO\textsubscript{3} using TIE, FlowTIE, and Gradient Descent.}
\label{tab:mse_by_thickness}
\end{table}
Table 1 reports the mean-squared phase error (MSE; lower is better) for GaAs and SrTiO$_3$ at two specimen thicknesses. 
FlowTIE attains the lowest error in three of the four settings and shows the largest gains for thicker samples. For GaAs at 5.7 Å, MSE is reduced by 49\%   and 44\% in comparison with TIE and Gradient Descent, respectively. At 28.3 Å\, the reduction is ~21\% against both baselines. In addition,
for SrTiO3 at 19.5 Å, the reduction is ~44\% versus both baselines.
 In all cases, an overall increase
in MSE with specimen thickness is observed, which is expected due to the increasing impact of dynamical
scattering effects at greater sample depths.
%\begin{table}[H]
%\centering
%\begin{tabular}{lccccc}
%\toprule
%\textbf{Material} & \textbf{Thickness (Å)} & \textbf{MSE (TIE)} & \textbf{MSE (FlowTIE)} & \textbf{MSE (Gradient Descent)} \\
%\midrule
%GaAs             & 5.7   & 0.0075 & 0.0038 & 0.0068 \\
%GaAs             & 28.3  & 0.1485 & 0.1167 & 0.1482 \\
%SrTiO\textsubscript{3} & 3.9   & 0.0073 & 0.0075 & 0.0072 \\
%SrTiO\textsubscript{3} & 19.5  & 0.1520 & 0.0853 & 0.1517 \\
%\midrule
%\multicolumn{5}{c}{
%\begin{tabular}{lccc}
%\textbf{Material} & \textbf{TIE (s)} & \textbf{FlowTIE (s)} & \textbf{Gradient Descent (s)} \\
%SrTiO\textsubscript{3} & 0.4622 ± 0.0769 & 0.4725 ± 0.1428 & 7.4761 ± 0.1862 \\
%GaAs                  & 0.5166 ± 0.1380 & 0.4864 ± 0.2424 & 8.0212 ± 0.8219 \\
%\end{tabular}
%} \\
%\bottomrule
%\end{tabular}
%\caption{Top: Phase estimation error (MSE) at different thicknesses. Bottom: Average computation time (in seconds) for GaAs and SrTiO\textsubscript{3} using TIE, FlowTIE, and Gradient Descent.}
%\label{tab:merged_top_bottom}
%\end{table}
\paragraph{Computational Time} We also report wall-clock runtimes measured on  GPU–equipped machine. The Gradient Descent baseline is implemented in PyTorch using automatic differentiation.  Table \ref{tab:avg_times} shows that TIE and FlowTIE have similar computation times (well under 1 s) in both GaAs and SrTiO$_3$, whereas Gradient Descent is an order of magnitude slower, taking several seconds per reconstruction as iterative process is performed.  These results highlight
a trade-off between accuracy and computational efficiency, where FlowTIE improves accuracy over TIE at a modest additional cost, while Gradient Descent incurs substantially longer processing times with limited accuracy benefits in our settings. 
% --- Table 2: Average computation time ---
\begin{table}[H]
\centering
\captionsetup{skip=10pt}
\begin{tabular}{lccc}
\toprule
\textbf{Material} & \textbf{TIE (s)} & \textbf{FlowTIE (s)} & \textbf{Gradient Descent (s)} \\
\midrule
SrTiO\textsubscript{3} & 0.4622 ± 0.0769 & 0.4725 ± 0.1428 & 7.4761 ± 0.1862 \\
GaAs                   & 0.5166 ± 0.1380 & 0.4864 ± 0.2424 & 8.0212 ± 0.8219 \\
\bottomrule
\end{tabular}
\caption{Average computation time (in seconds) for GaAs and SrTiO\textsubscript{3} using TIE, FlowTIE, and Gradient Descent.}
\label{tab:avg_times}
\end{table}

\section{Conclusion and Future Work}
This study presents an initial demonstration of applying the Transport of Intensity Equation (TIE) within 4D-STEM framework and integrating it with a flow-based generative model. By combining the physical constraints of TIE, in terms of continuity equation, and neural network model to learn vector field, FlowTIE offers a hybrid approach that blends physics-based modeling with data-driven learning. The use of a simple convolutional architecture reflects the exploratory nature of this work, leaving room for improvement through more advanced models.  Future research will explore transformer-based or attention-driven architectures to better capture complex phase behavior, especially in the scenario with strong dynamical scattering effect. Furthermore, we will evaluate the method on experimental data, both in- and out-of-distribution, to assess performance under realistic conditions. Additionally, we will conduct systematic hyperparameter fine-tuning and ablation studies to quantify the trade-off between physics-based losses and data fidelity. Although this work focuses on 4D-STEM phase reconstruction, the framework is broadly applicable to other microscopy modalities.
\newpage
\section*{Acknowledgment}
 The authors gratefully acknowledge computing time on the supercomputer JURECA\cite{thornig2021jureca} at Forschungszentrum Jülich under grant  \texttt{delia-mp}.

\bibliographystyle{ieeetr}
\bibliography{references}
 
\end{document}